\documentclass[10pt,twocolumn,letterpaper]{article}

\usepackage[pagenumbers]{cvpr} % To force page numbers, e.g. for an arXiv version

\usepackage{graphicx}
\usepackage{amsmath}
\usepackage{amssymb}
\usepackage{booktabs}
\usepackage{multirow}

\usepackage[pagebackref,breaklinks,colorlinks]{hyperref}
\usepackage[english]{babel}
\usepackage{amsthm}
\usepackage{pifont}% http://ctan.org/pkg/pifont
\usepackage{color, colortbl}

\usepackage[capitalize]{cleveref}
\crefname{section}{Sec.}{Secs.}
\Crefname{section}{Section}{Sections}
\Crefname{table}{Table}{Tables}
\crefname{table}{Tab.}{Tabs.}

\def\cvprPaperID{2349} % *** Enter the CVPR Paper ID here
\def\confName{CVPR}
\def\confYear{2023}

\begin{document}
% %%%%%%% Definition %%%%%%%%
\theoremstyle{definition}
\newtheorem{definition}{Definition}[section]
\theoremstyle{remark}
\newtheorem*{remark}{Remark}
\newcommand{\cmark}{\ding{51}}%
\newcommand{\xmark}{\ding{55}}%
\definecolor{Gray}{gray}{0.9}

\title{DATE: Dual Assignment for End-to-End Fully Convolutional Object Detection}
% %%%%%%%%%%%%%%%%%%%%%%%%%% Title Candidates %%%%%%%%%%%%%%%%%%%%%%%%%%%%
% 1. Free Lunch for End-to-End Fully Convolutional Object Detection
% 2. Dual Assignment for End-to-End Fully Convolutional Object Detection
% %%%%%%%%%%%%%%%%%%%%%%%%%% Title Candidates %%%%%%%%%%%%%%%%%%%%%%%%%%%%

\author{Yiqun Chen$^1$, Qiang Chen$^2$, Qinghao Hu$^1$, Jian Cheng$^1$\\
$^1$Institute of Automation, Chinese Academy of Sciences \quad $^2$Baidu VIS\\
{\tt\small \{chenyiqun2021, huqinghao2014\}@ia.ac.cn, chenqiang13@baidu.com, jcheng@nlpr.ia.ac.cn}
% \and
}
\maketitle

%%%%%%%%% ABSTRACT
% \begin{abstract}
% \textcolor{blue}{Fully convolutional detectors discard one-to-many assignment and adopt one-to-one assignment to achieve end-to-end detection but suffer from the slow convergence issue. In this paper, we revisit these two assignment methods. We find that bringing one-to-many assignment back to end-to-end fully convolutional detectors helps with the model convergence. Based on the finding, we propose {\em Dual Assignment} for end-to-end fully convolutional detection. Our method constructs two branches with one-to-many and one-to-one assignment during training. It only uses the branch with one-to-one assignment for model inference. The experiments show that Dual Assignment gives nontrivial improvements and speeds up model convergence upon OneNet and DeFCN. Moreover, we provide a hypothesis on the joint training of two branches, which offers an intuitive understanding of the results and provides guidelines for further exploration.}
% \end{abstract}
\begin{abstract}
   Fully convolutional detectors discard the one-to-many assignment and adopt a one-to-one assigning strategy to achieve end-to-end detection but suffer from the slow convergence issue. In this paper, we revisit these two assignment methods and find that bringing one-to-many assignment back to end-to-end fully convolutional detectors helps with model convergence. Based on this observation, we propose {\em \textbf{D}ual \textbf{A}ssignment} for end-to-end fully convolutional de\textbf{TE}ction (DATE). Our method constructs two branches with one-to-many and one-to-one assignment during training and speeds up the convergence of the one-to-one assignment branch by providing more supervision signals. DATE only uses the branch with the one-to-one matching strategy for model inference, which doesn't bring inference overhead. Experimental results show that Dual Assignment gives nontrivial improvements and speeds up model convergence upon OneNet and DeFCN. Code: {https://github.com/YiqunChen1999/date}. 
%   \textcolor{red}{Moreover, we provide a hypothesis on the joint training of two branches, which offers an intuitive understanding of the results and provides guidelines for further exploration.}
\end{abstract}

%%%%%%%%% BODY TEXT
\section{Introduction}
\label{sec:intro}
\begin{figure}[ht!]
    \centering
    \includegraphics[trim={4.5cm 3cm 4.5cm 3cm}, clip, width=0.47\textwidth]{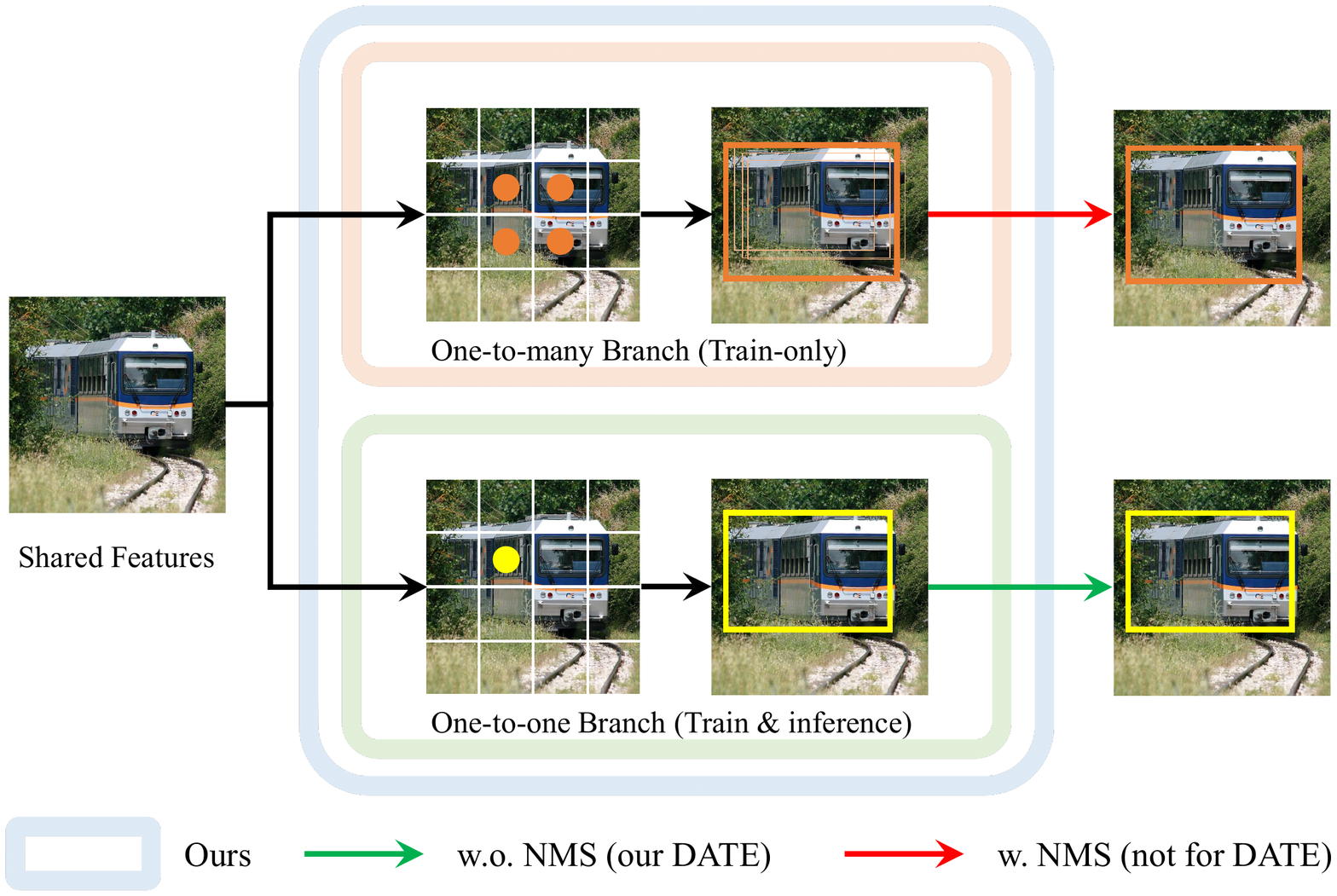}
    \caption{We jointly train one-to-one and one-to-many branches and only keep the former one during inference. Typical detectors are trained with the one-to-many assignment (the top pathway), \eg, FCOS, and require non-maximum suppression (NMS) as post-processing to remove duplicated predictions during inference. The one-to-many assignment strategy makes the probability and bounding boxes of selected samples (orange boxes) as close to ground truth as possible. OneNet replaces the one-to-many assignment with a one-to-one assignment strategy and realizes end-to-end detection without anchor generation and NMS, see the bottom pathway. The one-to-one matching only selects one sample (yellow box) based on the matching quality for one ground truth. Bounding boxes and the selected points are for demonstration only and are not produced by code. Best viewed in color.}
    \label{fig:core}
\end{figure}

One-stage convolutional object detectors, \eg, RetinaNet~\cite{RetinaNet-2017} and FCOS~\cite{FCOS-2019}, are widely adopted by the community for their simplicity. Despite their success, the one-to-many assignment (o2m) strategy makes them rely on non-maximum suppression (NMS) to remove duplicated predictions. Such a process makes them sensitive to the hyperparameter of NMS and may lead to a sub-optimal solution. 

This problem motivates researchers to remove the non-maximum suppression for realizing end-to-end fully convolutional object detection. Inspired by DETR~\cite{DETR-2020}, OneNet~\cite{OneNet-2021} discusses what makes for end-to-end detection. By conducting detailed experiments on Hungarian matching~\cite{Hungarian-1955}, OneNet recognizes that the one-to-one assignment (o2o) strategy with classification cost is the key to end-to-end detection. Specifically, their results suggest that assigning only one prediction for a ground truth by considering classification cost prevents it from producing redundant predictions. 

Despite being simple and realizing the end-to-end detection, OneNet suffers from a slow convergence issue. Our explorations show that OneNet requires more training time than its counterpart FCOS to achieve competitive performance. As a result, the model trained with the one-to-one assignment is inferior to the one with the one-to-many assigning strategy under the same setting~\footnote{12 epochs: 35.4 AP (OneNet) \vs 37.2 (FCOS) on COCO~\cite{COCO-2014}}. However, as mentioned above, the one-to-many positive samples assignment strategy makes the NMS necessary in removing duplicated predictions during inference. Based on these observations, it's natural to ask: {\textit{Can we train an end-to-end detector with competitive performance under the same setting? }}

In this paper, our answer is yes. {We suppose the reason for the slower convergence of OneNet is the weaker supervision for the feature extractor from the one-to-one assignment strategy. Specifically, the one-to-one assignment policy provides fewer positive samples than the one-to-many assignment, resulting in the lack of classification and regression supervisory signals of feature extractors.} The above findings motivate us to combine the advantages of one-to-one and one-to-many assignment strategies. We re-introduce the one-to-many assignment strategy to OneNet and propose a {\textbf{D}ual \textbf{A}ssignment for de\textbf{TE}ction (DATE)} to solve this problem. Specifically, as illustrated in \cref{fig:core}, we jointly train the one-to-many assignment and the one-to-one assignment branches. Once we finish the training, we only keep the predictor trained with the one-to-one positive sample assignment to realize end-to-end detection. 

Experimental results suggest that our dual assignment strategy speeds up the convergence of the end-to-end detector. Our NMS-free DATE can surpass or be on par with its NMS-based counterpart after 12 and 36 epochs of training~\footnote{12 epochs: 37.3 AP (Ours) \vs 37.2 AP (FCOS); 36 epochs: 40.6 AP (Ours) \vs 39.8 AP (FCOS)}. Thanks to the lightweight one-to-many matching branch, \eg, only two or three convolutional layers, we cost little extra computational resources during training but enjoy significant performance improvement in model inference. 

{Our contributions can be summarized as follows: \begin{itemize}
    \item We propose a Dual Assignment strategy to speed up the convergence of end-to-end fully convolutional detectors by introducing more supervision signals.
    \item The proposed Dual Assignment policy introduces negligible cost during training and no overhead during inference.
    \item Based on the proposed Dual Assignment strategy, our simple yet effective DATE achieves competitive or slightly better performance compared with models based on the one-to-many assignment policy.
\end{itemize}
}

% =========================== Related Works =============================
\section{Related Works}
\label{sec:related}
The works most related to our method are fully convolutional detection and end-to-end detection. 

\noindent \textbf{Fully Convolutional Detection.} 
Many approaches in the literature explored fully convolutional detection. 
Typically an object detector, \eg, a one-stage~\cite{RetinaNet-2017, YOLOv2-2017, YOLOv4-2020, SSD-2016} or two-stage~\cite{FasterRCNN-2015, CascadeRCNN-2018} one, requires anchor box generation to generate numerous pre-defined boxes for each location. These anchor boxes serve as prior knowledge for bounding boxes regression. These boxes need carefully tuned hyperparameters to define the size and shape. CornerNet~\cite{CornerNet-2018} and FCOS~\cite{FCOS-2019} got rid of anchor box generation and realized anchor-free object detection. The former one detects two sets of corner points and then matches them. In this paper, we prefer FCOS for its simplicity, which predicts a bounding box from a location. However, anchor-free methods~\cite{YOLO-2016, CenterNet-2019, ATSS-2020, RepPointsV2-2020, AutoAssign-2020, GeneralizedFocalLoss-2020, TOOD-2021, Foveabox-2020, Reppoints-2019} still rely on NMS as post-processing to remove duplicated predictions. In this work, we eliminate the NMS without modifying the computing process. 

\noindent \textbf{End-to-End Detection.} 
Recently, the community has made many efforts to achieve end-to-end detection. 
Learnable NMS~\cite{LearnableNMS-2017} and Relation Networks~\cite{RelationNetworks-2018} builds a deep network for a convolutional detector to filter out duplicated predictions. Such a method introduces extra parameters and computing costs. Instead of modifying the NMS, methods~\cite{LSTM-E2E-2016, RNN-Det-2015} based on the recurrent neural network (\eg, LSTM~\cite{LSTM-1997}) generate a set of bounding boxes directly. Despite removing the NMS, they are considered limited to some specific fields. Transformer-based~\cite{Transformer-2017} DETR~\cite{DETR-2020} and its variants~\cite{DeformableDETR-2020, PnPDETR-2021, ConditionalDETR-2021, DN-DETR-2022, EfficientDETR-2021, UP-DETR-2021, SAM-DETR-2022} share the same idea of set prediction~\cite{DeepSetNet-2017, SetPrediction-2019} with RNN-based models. DETR-like models typically learn a set of object queries as ``feature probes'' to detect objects. Recently $\mathcal{H}$-DETR~\cite{H-DETR-2022} proposed a hybrid matching strategy to speed up the convergence of DETR-like models. However, due to the unique object queries of DETR, it's hard to apply the hybrid matching policy to convolutional detectors. 
In this paper, we manage to realize end-to-end fully convolutional detection with a simple design. 
% =========================== Method =============================
\section{Method}
\label{sec:method}
\begin{figure*}[t!]
    \centering
    \includegraphics[trim={2.8cm 7.0cm 2.8cm 7.4cm}, clip, width=\textwidth]{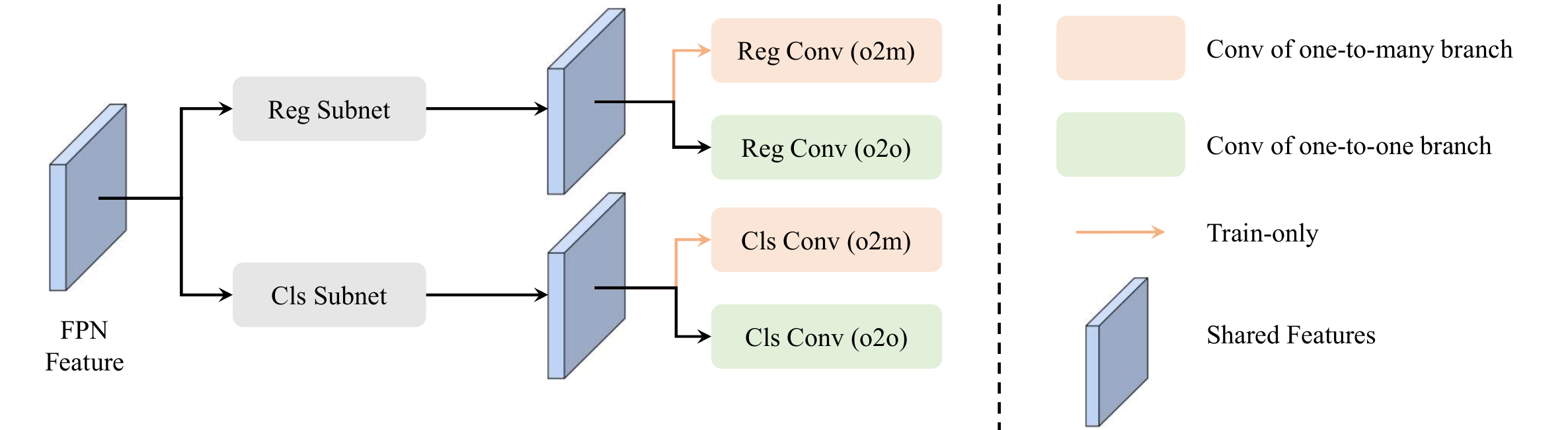}
    \caption{Diagram of the two branches (RetinaNet-style one-to-many assignment branch). Each branch only consists of two convolutional layers, \ie, one for classification and the other for regression. When using the FCOS-style one-to-many assignment branch, there will be an additional convolutional layer for center-ness. This layer is not shown in the figure. }
    \label{fig:method-arch}
\end{figure*}
To overcome the shortcoming of lacking supervision signals, \ie, few positive samples, of the one-to-one assignment, we propose the \textbf{D}ual \textbf{A}ssignment for end-to-end fully convolutional de\textbf{TE}ctors (DATE), as shown in \cref{fig:method-arch}. Following typical one-stage detectors, our architecture takes as input an image and extracts multi-scale features for classification and regression. Our multi-scale feature extractor consists of a backbone~\cite{ResNet-2016}, a feature pyramid network~\cite{FPN-2017}, and two subnets~\cite{RetinaNet-2017}. {Our Dual Assignment then constructs two branches ($B_{o2o}$ and $B_{o2m}$) following the feature extractor during training and only keeps one branch for end-to-end detection during inference, described next. }

\subsection{Dual Assignment}\label{sec:method-arch}

A significant concept of our method is the Dual Assignment. An assignment policy assigns one ground-truth to one or many predictions to supervise the network. Our dual assignment strategy constructs \textit{two branches} (described latter) with different sample assignment strategies during training. The first branch, denoted by $B_{o2o}$, is trained with a one-to-one assignment strategy (\cref{fig:core}, bottom). The second one, named $B_{o2m}$, adopts a one-to-many assignment policy (\cref{fig:core}, top) during training. These branches take as input the shared classification features and the regression features to make predictions. Then their assignment policy will construct (ground-truth, prediction) pairs to calculate the losses. 

Optimizing our DATE is a multi-objective optimization problem.
% \begin{align}
%     \min l(x) &= \begin{bmatrix}
%     l_{o2o}(x), l_{o2m}(x)
%     \end{bmatrix}, \quad x \in \mathcal{S}
% \end{align}
% where $\mathcal{S}$ is the \textit{feasible design space}~\footnote{also called the \textit{feasible set}}, $l_{o2o}$ and $l_{o2m}$ are the losses of the one-to-one and the one-to-many assignment branch, respectively. 
Ideally, we would like to seek a \textit{utopia solution}~\cite{ParetoOptimality-2017} (see supplementary materials) that minimizes losses of both branches simultaneously. However, a solution can hardly be the local minima of objective functions of different tasks at the same time. Attaining a \textit{Pareto optimal solution}~\cite{ParetoOptimality-2017} (also see supplementary materials) of the Pareto frontier is a more common choice. 

Since there are infinite Pareto solutions for the multi-objective optimization problem, we often need to make decisions to choose one or some from them. Ideally, the optimization process should reflect the preferences of users. One of the most common approaches is the weight sum method, which assigns the weights of each objective function: 
\begin{align} \label{eq:loss-date}
    l_{DA} = \lambda_{o2o} l_{o2o} + \lambda_{o2m} l_{o2m} 
\end{align}
where $l_{o2o}$ and $l_{o2m}$ are the losses of the one-to-one and the one-to-many assignment branch, respectively. $\lambda_{o2o}, \lambda_{o2m} \ge 0$ are the weights of $l_{o2o}$ and $l_{o2m}$, respectively. Intuitively, a relatively larger weight than others means the corresponding objective function is more important than the rest. For instance, in an extreme case, if we set $\lambda_{o2m}$ in \cref{eq:loss-date} as zero, it will degenerate into training only the one-to-one branch without caring about the one-to-many branch, and vice versa. 

\noindent \textbf{One-to-one Assignment Branch.} As illustrated in \cref{fig:method-arch}, this branch called $B_{o2o}$ consists of two convolutional layers, one for classification and the other for regression. The one-to-one assignment strategy will assign one ground-truth to one prediction and construct $G$ (ground-truth, prediction) pairs, where $G$ is the number of the annotated objects in one image. Predictions not matched to any ground truth are negative samples. Typically, the community adopts the Hungarian matching~\cite{Hungarian-1955, DETR-2020, OneNet-2021} with cost or quality measurement as the one-to-one assignment algorithm. We train this branch with the one-to-one assignment by following OneNet but use POTO~\cite{DeFCN-2021} as quality measurement. We supervise this branch via:
\begin{align}
    l_{o2o} = \alpha_{cls} l_{o2o}^{cls} + \alpha_{reg} l_{o2o}^{reg} + \alpha_{iou} l_{o2o}^{iou}
\end{align}
where $l_{o2o}^{cls}$, $l_{o2o}^{reg}$ and $l_{o2o}^{iou}$ are classification, regression and IoU loss~\cite{RetinaNet-2017, FastRCNN-2015, GIoU-2019}, respectively. $\alpha_{cls}$, $\alpha_{reg}$ and $\alpha_{iou}$ are the weights of the corresponding loss. We have these weights the same as OneNet. 

\noindent \textbf{One-to-many Assignment Branch.} The one-to-many assignment assigns one ground-truth to several predictions and constructs $N$ (ground-truth, prediction) pairs, \eg, max IoU assignment strategy. Typically $N > G$. We construct this branch ($B_{o2m}$) by mainly considering the widely studied one-stage detectors RetinaNet~\cite{RetinaNet-2017} and FCOS~\cite{FCOS-2019}. Specifically, our DATE adopts the two convolutional layers of RetinaNet or three layers of FCOS. We train this branch by minimizing:
\begin{align}
    l_{o2m} = \beta_{cls} l_{o2m}^{cls} + \beta_{reg} l_{o2m}^{reg} + \beta_{ctr} l_{o2m}^{ctr}
\end{align}
where $\beta_{ctr}$ and $l_{o2m}^{ctr}$ are the weight and loss function of center-ness prediction of FCOS. The meaning of other symbols is analogous to the one-to-one assignment branch. $\beta_{ctr}$ is 0 for RetinaNet as there is no center-ness prediction in RetinaNet. Once we finish the training, we will discard this branch. Weights are kept the same as RetinaNet or FCOS. 

\subsection{Discussion} \label{sec:method-discussion}

The one-to-one assignment only provides positive samples with the same number as the ground-truth, which is less than the one-to-many assignment policy. We suppose the slow convergence issue of the detectors trained with the one-to-one assignment strategy happens mainly due to insufficient supervision signals. Because of the lacking of sufficient positive samples, the feature extractor may need more iterations of training to produce suitable features for classification and regression. 

In contrast, the one-to-many assignment strategy assigns multiple positive samples to a ground-truth. We hypothesize that a sufficient number of positive samples reduce the necessity for more iterations of training. However, the one-to-many assignment policy relies on the NMS  to remove the duplicated predictions. 

Our Dual Assignment combines the advantages of the one-to-one and the one-to-many assignment. The one-to-many assignment branch $B_{o2m}$ provides sufficient supervision signals to speed up the training of the shared multi-scale feature extractor. The one-to-one assignment branch acts like receiving and tuning the features to make end-to-end detection. We guess the one-to-many assignment strategy helps relieve the optimizing issue, making the one-to-one assignment branch focus more on fitting the received representations. 
{In \cref{sec:results-dual-assignment}, we empirically show that the features extractor supervised by the one-to-many branch is a significant factor in speeding up the convergence of DATE. }

Training our DATE is like training two networks with parameter sharing, then discarding one of them (\eg, the one-to-many assignment branch) during inference. This design does not introduce any overhead during model inference (see \cref{sec:results-superiority}). Due to the shared parameters, our DATE only introduces negligible cost during training, making it friendly to the resources limited machines. The proposed Dual Assignment makes our method orthogonal to the modification for the one-to-one assignment branch, making it possible to integrate other improvements. We show in \cref{sec:results-defcn} that improving the one-to-one assignment branch can further boost performance. 

% =========================== Experimental Results =============================
\section{Experimental Results}
\label{sec:results}

\begin{table*}[t!]
    \centering
    \begin{tabular}{l|c|c|llllll|cc}
        \toprule
        Model        & epoch & NMS
        & $\text{AP}$ & $\text{AP}_{50}$ & $\text{AP}_{75}$ & $\text{AP}_s$ & $\text{AP}_m$ & $\text{AP}_l$ & Params (M) & GFLOPs\\
        \midrule

        % 0.369 0.567 0.390 0.207 0.402 0.498
        RetinaNet & 12 & \cmark 
        & 36.9 & 56.7 & 39.0 & 20.7 & 40.2 & \textbf{49.8} 
        & 33.6 & 233\\

        % 37.2 56.8 39.6 20.6 40.5 49.8
        FCOS & 12  & \cmark
        & 37.2 & \textbf{56.8} & 39.6 & 20.6 & \textbf{40.5} & \textbf{49.8} 
        & 32.0 & 201\\
        
        % 0.354 0.533 0.381 0.193 0.386 0.454
        OneNet  & 12  & \xmark
        & 35.4 & 53.3 & 38.1 & 19.3 & 38.6 & 45.4 
        & 32.0 & 201\\

        \midrule

        % 0.373 0.553 0.407 0.212 0.403 0.488
        DATE-F (Ours) & 12  & \xmark
        & \textbf{37.3 (+1.9)} & 55.3 & \textbf{40.7} & \textbf{21.2} & 40.3 & 48.8 
        & 32.0 & 201\\

        \midrule \midrule

        % 0.398 0.597 0.423 0.252 0.433 0.512
        FCOS & 36  & \cmark
        & 39.8 & \textbf{59.7} & 42.3 & 25.2 & 43.3 & \textbf{51.2} 
        & 32.0 & 201\\

        % 0.386 0.566 0.422 0.239 0.417 0.487 (w.o. NMS)
        % 0.387 0.586 0.414 0.235 0.414 0.495 (w.   NMS)
        OneNet & 36  & \xmark
        & 38.6 & 56.6 & 42.2 & 23.9 & 41.7 & 48.7 
        & 32.0 & 201\\

        \midrule

        % 0.406 0.589 0.444 0.256 0.441 0.509 (w.o. NMS)
        DATE-F (Ours) & 36  & \xmark
        & \textbf{40.6 (+2.0)} & 58.9 & \textbf{44.4} & \textbf{25.6} & \textbf{44.1} & 50.9 
        & 32.0 & 201\\
        
        \bottomrule

    \end{tabular}
    \caption{The OneNet, which is trained with the one-to-one assignment, is inferior to models trained with the one-to-many assignment. Our Dual Assignment relieve the slow convergence issue and reports competitive or slightly better results than detectors trained with the one-to-many assignment. Numbers in the brackets are improvements over OneNet. We don't calculate the FLOPs of the NMS. }
    \label{tab:results-observation}
\end{table*}

In this section, we present detailed experiments of our {\textbf{D}ual \textbf{A}ssignment for de\textbf{TE}ction (DATE)}. 
% We pay special attention to the characteristics as well as promising extensions of our method. 
More details are available in supplementary materials.
\subsection{Baseline Settings}
Unless specified, we use the following settings for our experiments. We conduct our experiments based on PyTorch~\cite{PyTorch-2019} and mmdetection~\cite{mmdetection-2019}. 

\noindent \textbf{Baseline.} We choose OneNet~\cite{OneNet-2021} as our baseline for its simplicity. This decision allows researchers to explore other improvements and promising extensions. We give an example in \cref{sec:results-defcn}. 

\noindent \textbf{Optimizer.} Following OneNet, we use AdamW~\cite{AdamW-2017} as our optimizer and set the learning rate as 4e-4 by default. Other hyperparameters, \eg, batch size and weight decay, are kept the same as with OneNet. 

\noindent \textbf{Training Schedules.} We conduct experiments with training schedules of 12 and 36 epochs. Other researchers also refer to these settings as 1x and 3x training settings in the literature~\cite{DN-DETR-2022, mmdetection-2019}. 

\noindent \textbf{One-to-many Branch.} We study the influence of different models as the one-to-many assignment (o2m) branches. ResNet~\cite{RetinaNet-2017} and FCOS~\cite{FCOS-2019} are the most widely studied one-stage detectors, so we mainly consider them in the experiments. Unless specified, only the two or three convolutional layers for predicting are not shared. To balance the loss, we set $\lambda_{o2m}$ as 1 and 2 when adopting FCOS and RetinaNet as $B_{o2m}$, respectively. 

\noindent \textbf{Notations.} To save space, we use the following notations: \begin{itemize}
    \item DATE-F: Our DATE trained with FCOS-style one-to-many assignment branch (three conv layers);
    \item DATE-R: Our DATE trained with RetinaNet-style one-to-many assignment branch (two conv layers);
\end{itemize}

\noindent \textbf{Datasets.} We mainly consider the following two challenging benchmarks: \begin{itemize}
    \item COCO Dataset: We mainly conduct experiments on the challenging dataset COCO~\cite{COCO-2014}. We train our model on the train2017 split (about 118k images) and evaluate the performance with the val2017 (5k images). Models are evaluated on the standard mean Average Precision defined by the COCO. 
    \item CrowdHuman: CrowdHuman~\cite{CrowdHuman-2018} is a challenging benchmark to evaluate the ability of crowded scene detection of detectors, which contains about 15k training images and 4k images for evaluation. Annotation boxes are highly crowded and overlapped. We follow OneNet to use AP, mMR, and recall to evaluate our model. 
\end{itemize}

\subsection{Superiority of DATE} \label{sec:results-superiority}

As shown in \cref{tab:results-observation}, the proposed NMS-free DATE is on par with the NMS-based methods, \eg, RetinaNet and FCOS. Our DATE reports 37.3 AP and 37.0 AP when adopting FCOS and RetinaNet as the o2m branch, respectively. These results are competitive compared with FCOS (37.2 AP) and RetinaNet (36.9 AP) trained under the same setting. 
Despite the competitive performance, our DATE does not rely on the NMS, while FCOS and RetinaNet require NMS as post-processing to remove duplicated predictions. 

Our DATE realizes the end-to-end detection and is superior to the OneNet with the same computation resources. After 12 epochs of training, the proposed DATE achieves 37.3 AP, which is 1.9 AP higher than the OneNet (35.4 AP). A similar phenomenon happens after 36 epochs of training, \eg, 40.6 AP of our DATE is better than 38.6 AP of OneNet. 
% Interestingly, changing the o2m branch from FCOS to RetinaNet does not benefit the one-to-one assignment branch, \eg, AP decreased from 37.3 to 37.0. %We explain why in the following.

Moreover, our DATE does not cost additional computing resources during inference but enjoys extra performance improvement, as shown in \cref{tab:results-observation}. We discard the one-to-many branch once the training is done and don't introduce additional parameters and FLOPs. The total parameters and the FLOPs of our DATE are the same as OneNet and FCOS, \ie, 32 M total parameters and 201 GFLOPs during inference. These advantages make our DATE a strong baseline. 

\begin{table*}[t!]
    \centering
    \begin{tabular}{l|c|c|llllll}
        \toprule
        Model       & settings & 3DMF
        & $\text{AP}$ & $\text{AP}_{50}$ & $\text{AP}_{75}$ & $\text{AP}_s$ & $\text{AP}_m$ & $\text{AP}_l$ \\ % & Params (M) & GLOPs\\
        \midrule

        % 37.8486,55.5946,41.7789,22.1394,41.3476,48.6532
        DeFCN & 1x & \cmark
        & 37.8 & 55.6 & 41.8 & 22.1 & 41.3 & 48.7 \\ % & 32.6 & 213 \\

        % 0.373 0.553 0.407 0.212 0.403 0.488
        DATE-F (Ours) & 1x & \xmark
        & 37.3 & 55.3 & 40.7 & 21.2 & 40.3 & 48.8 \\ % & 32.0 & 201 \\

        % 0.389 0.571 0.429 0.225 0.421 0.513
        DATE-F (Ours) & 1x & \cmark
        & \textbf{38.9(+1.1)} & \textbf{57.1(+1.4)} & \textbf{42.9(+1.1)} & \textbf{22.5(+0.4)} & \textbf{42.1(+0.8)} & \textbf{51.3(+2.4)} \\ % & 32.6 & 213 \\

        \hline 

        % | 41.443 | 59.520 | 45.650 | 26.075 | 44.914 | 52.025 |
        DeFCN & 3x & \cmark
        & 41.4 & 59.5 & 45.7 & 26.1 & 44.9 & 52.0 \\ % & 32.6 & 213 \\

        % 0.373 0.553 0.407 0.212 0.403 0.488
        DATE-F (Ours) & 3x & \xmark
        & 40.6 & 58.9 & 44.4 & 25.6 & 44.1 & 50.9 \\ % & 32.0 & 201 \\

        % 0.420 0.603 0.462 0.273 0.455 0.530
        DATE-F (Ours) & 3x & \cmark
        & \textbf{42.0(+0.6)} & \textbf{60.3(+0.8)} & \textbf{46.2(+0.5)} & \textbf{27.3(+0.2)} & \textbf{45.5(+0.6)} & \textbf{53.0(+1.0)} \\ % & 32.6 & 213 \\

        \bottomrule
    \end{tabular}
    \caption{Comparison with DeFCN. To make a fair comparison, we add the 3D Max Filtering (3DMF) of DeFCN. Numbers in the brackets are the improvements over DeFCN. 1x means 90k iterations (DeFCN) or 12 epochs (ours, about 88k iterations) and 3x means 270k iterations or 36 epochs (about 264k iterations).}
    \label{tab:results-defcn}
\end{table*}
\subsection{Promising Extension to DATE}\label{sec:results-defcn}

% \noindent \textbf{Improving on One-to-one Assignment Branch.} 
As mentioned in \cref{sec:method-discussion}, our DATE is orthogonal to the improvements to the network, which make it possible to integrate new components. Recently an end-to-end detector DeFCN~\cite{DeFCN-2021} proposed to modify the network to improve its performance. Besides using POTO as a quality measurement and adding an auxiliary loss, DeFCN also proposes a novel 3D Max Filtering (3DMF). Such a 3DMF module processes the multi-scale features to improve the discriminability of convolutions in the local region. 

We explore the influence of integrating new components and take DeFCN as an example. Results are available in \cref{tab:results-defcn}~\footnote{Results of DeFCN are produced by the released official code. }. Our pure DATE is inferior to the full version of DeFCN. By eliminating the difference, we report the results significantly better than DeFCN. Our DATE reports 38.9 AP after 12 epochs of training, which is 1.0 AP higher than the 37.9 AP of DeFCN. This phenomenon happens for 3x training schedules, but the difference decreases, \ie, DATE with 42.0 AP surpasses DeFCN with 41.4 AP.

Incorporating 3DMF is a case of extending our DATE to be a more powerful end-to-end detector. 
Thanks to the model and the computing during inference being as simple as OneNet, many promising improvements still unexplored. We hope the proposed method will serve as a strong baseline for these explorations. We believe that better modifications will bring higher performance. 

\begin{table}[t!]
    \centering
    \begin{tabular}{l|rrrrrr}
        \toprule
        Model & $\text{AP}$ & $\text{AP}_{50}$ & $\text{AP}_{75}$ & $\text{AP}_s$ & $\text{AP}_m$ & $\text{AP}_l$ \\
        \midrule
        
        % 0.396 0.579 0.431 0.237 0.429 0.496
        OneNet 
        & 39.6 & 57.9 & 43.1 & 23.7 & 42.9 & 49.6 \\
        
        %  0.411 0.617 0.434 0.251 0.448 0.539
        RetinaNet 
        & 41.1 & \textbf{61.7} & 43.4 & 25.1 & 44.8 & 53.9 \\
        
        % 0.416 0.615 0.443 0.257 0.456 0.530
        FCOS 
        & 41.6 & {61.5} & 44.3 & 25.7 & 45.6 & 53.0 \\
        
        % 0.408 0.608 0.435 0.256 0.442 0.519
        FCOS$\dagger$ 
        & 40.8 & 60.8 & 43.5 & 25.6 & 44.2 & 51.9 \\

        \midrule

        % 0.422 0.606 0.463 0.266 0.458 0.541
        DATE-F 
        & \textbf{42.2} & 60.6 & \textbf{46.3} & \textbf{26.6} & \textbf{45.8} & \textbf{54.1} \\
        
        % % 0.430 0.614 0.473 0.274 0.465 0.541
        % DATE+3DMF 
        % & 43.0 & 61.4 & 47.3 & 27.4 & 46.5 & 54.1 \\

        \bottomrule
    \end{tabular}
    \caption{Results when adopting ResNet-101 as backbone. $\dagger$ means the FCOS is the trained DATE-F but discards the one-to-one assignment branch instead of the one-to-many assignment branch.}
    \label{tab:results-stronger-backbone}
\end{table}
\subsection{Stronger Backbone}

To further demonstrate robustness and effectiveness, we explore the influence of a stronger backbone. Detailed results are available in \cref{tab:results-stronger-backbone}. When adopting ResNet-101~\cite{ResNet-2016} as the backbone, our DATE achieves 42.2 AP, which is 2.6 AP higher than our baseline OneNet (\ie, 39.6 AP). This result suggests that our Dual Assignment still works when introducing a stronger backbone. Recall that when using ResNet-50 as the backbone (results are available in \cref{tab:results-observation}), the performance gap between our DATE (40.6 AP) and the baseline OneNet (38.6 AP) is 2.0 AP. The stronger backbone enlarges the performance gap. 

We also compare the performance with FCOS. Similar to the results of ResNet-50 as the backbone (\cref{tab:results-observation}), our DATE (42.2 AP) is still slightly better than FCOS (41.6 AP). 
% These results imply that the one-to-one assignment branch still converges well, and finding a local minimum may not be the main problem for a plain OneNet. Our Dual Assignment helps relieve the optimization issue of OneNet and further approaches the local minima. 

\subsection{Crowded Scene Detection}

\begin{table}[t!]
    \centering
    \begin{tabular}{l|rrr}
        \toprule
        Method & $\text{AP}_{50}$ $\uparrow$ & mMR$\downarrow$ & Recall$\uparrow$ \\
        \midrule

        Annotations
        & - & - & 95.1 \\

        % AP: 0.8110, mMR: 0.6160, recall: 0.8712
        RetinaNet
        & 81.1 & 61.6 & 87.1 \\

        % AP: 0.8182, mMR: 0.5512, recall: 0.8675
        FCOS
        & 81.8 & 55.1 & 86.8 \\

        % AP: 0.9014, mMR: 0.5003, recall: 0.9792
        OneNet
        & 90.1 & 50.0 & \textbf{97.9} \\
        
        \midrule
        
        % AP: 0.9052, mMR: 0.4896, recall: 0.9791
        DATE-F (Ours)
        & 90.5 & 49.0 & \textbf{97.9} \\
        
         % AP: 0.9057, mMR: 0.4838, recall: 0.9794
        DATE-R (Ours)
        & \textbf{90.6} & \textbf{48.4} & \textbf{97.9} \\

        \bottomrule
    \end{tabular}
    \caption{Crowd detection on CrowdHuman dataset. We train all models for 30k iterations. $\downarrow$ means the lower, the better. Multi-scale data augmentation is applied. Recall of applying NMS on annotations is reported in~\cite{DeFCN-2021}.}
    \label{tab:results-crowdhuman}
\end{table}

Crowded scene detection is quite challenging because of the overlapped boxes. Detectors with NMS tend to remove some true positives because of the overlap. 

Results in \cref{tab:results-crowdhuman} suggest that the end-to-end detector performs better than its NMS-based counterpart on CrowdHuman~\cite{CrowdHuman-2018}. Besides, our method still works under crowd scene detection, \eg, reducing about 5\% relative error on $\text{AP}_{50}$ over OneNet. {In contrast, the NMS-based detectors, \ie, RetinaNet and FCOS, report 81.1 AP and 81.8 AP, which is much lower than the end-to-end detectors, \ie, OneNet (90.1 AP) and our DATE (90.5 AP and 90.6 AP). }

Applying NMS on the annotations only achieves a recall of 95.1~\cite{DeFCN-2021}, which is slower than the NMS-free methods, \eg, OneNet and our DATE. This phenomenon suggests that NMS-based detectors will be bounded by the recall, making them unsuitable for crowded scene detection. 

\subsection{Why Dual Assignment}\label{sec:results-dual-assignment}

\begin{figure}[t!]
    \centering
    \includegraphics[trim={0.5cm 6.5cm 1.5cm 7.0cm}, clip, width=0.47\textwidth]{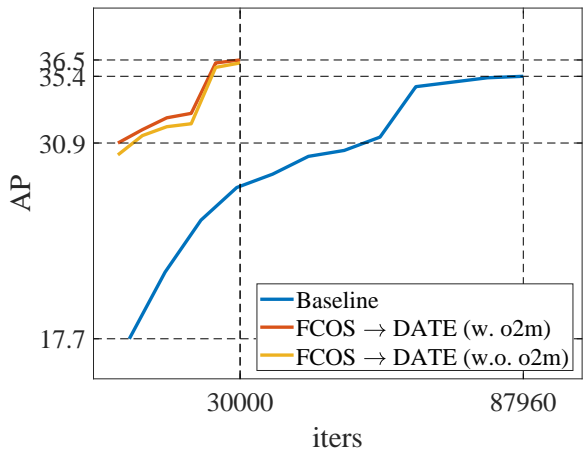}
    \caption{Initializing the multi-scale feature extractor with a trained FCOS. Our DATE can surpass the baseline in 30k iterations, which convergences significantly faster than the baseline (OneNet).}
    \label{fig:transferred}
\end{figure}
\begin{figure*}[t!]
    \centering
    \begin{subfigure}[b]{0.3\textwidth}
        \centering
        \includegraphics[trim={1cm 7.5cm 2.2cm 6cm}, clip, width=\textwidth]{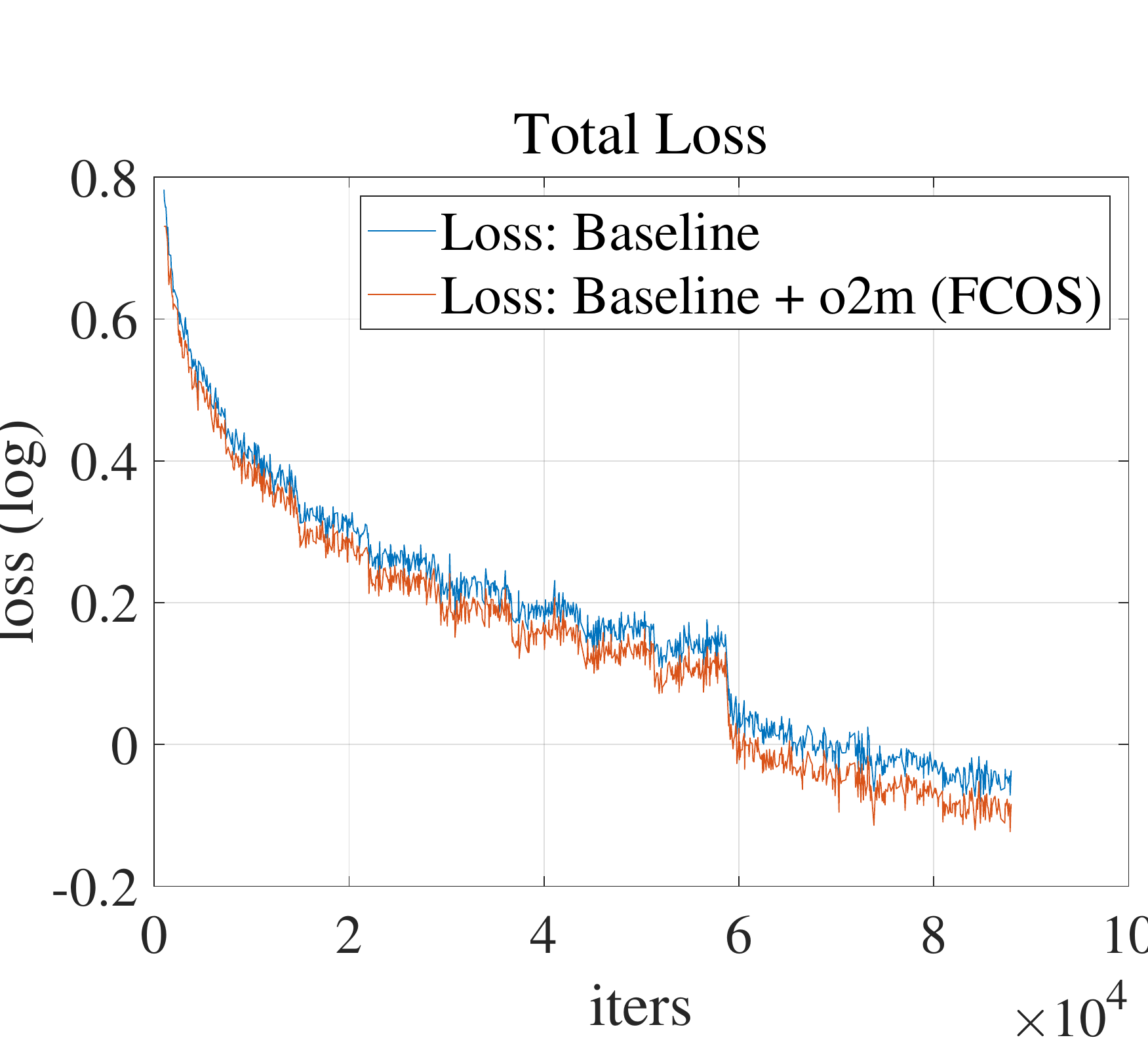}
        \caption{The total loss of the o2o branch. }
        \label{fig:results-dual-assignment-tot-loss}
     \end{subfigure}
     \hfill
    \begin{subfigure}[b]{0.3\textwidth}
        \centering
        \includegraphics[trim={1cm 7.5cm 2.2cm 6cm}, clip, width=\textwidth]{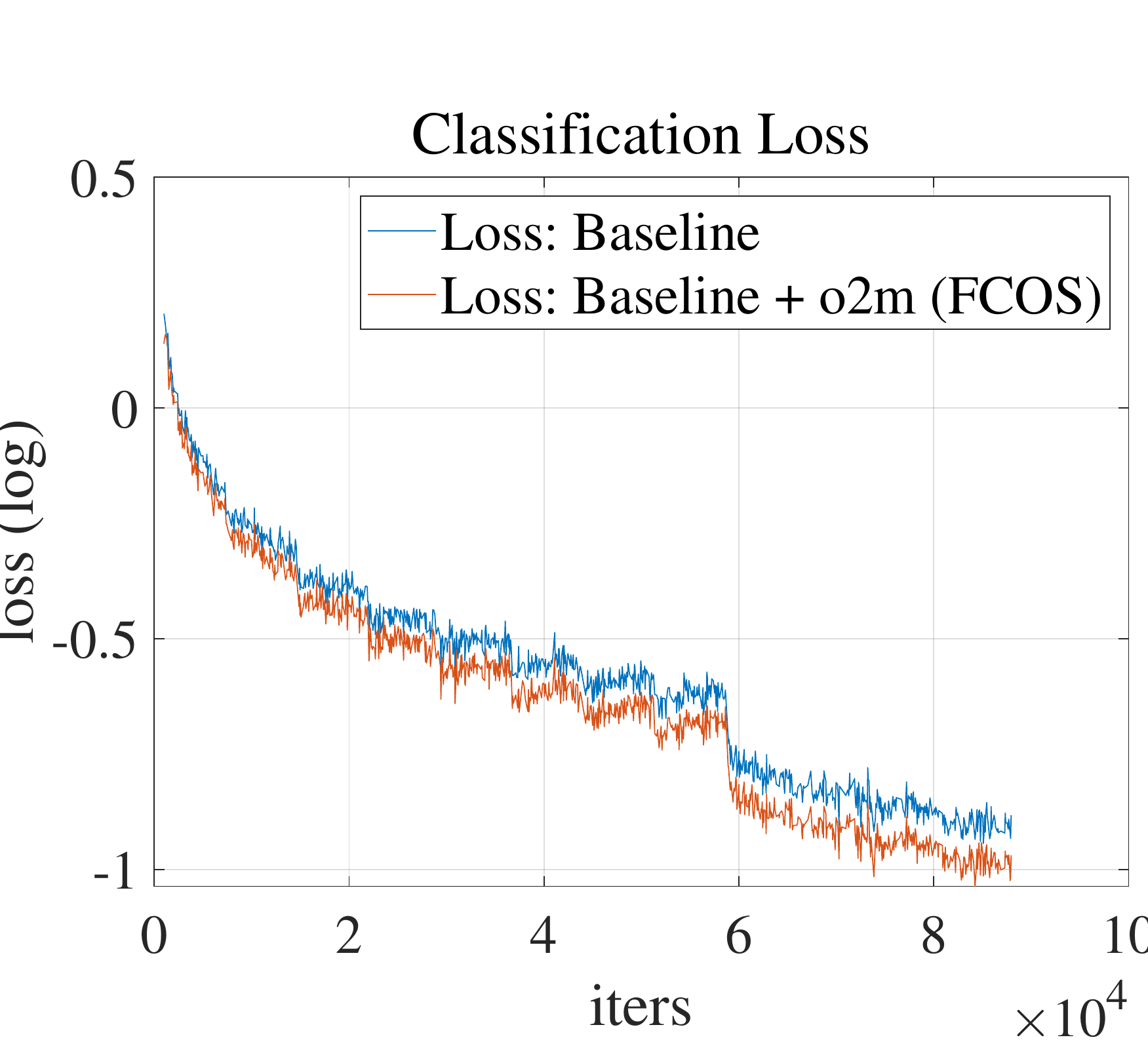}
        \caption{The classification loss of the o2o branch. }
        \label{fig:results-dual-assignment-cls-loss}
     \end{subfigure}
     \hfill
     \begin{subfigure}[b]{0.3\textwidth}
        \centering
        \includegraphics[trim={1cm 7.5cm 2.2cm 6cm}, clip, width=\textwidth]{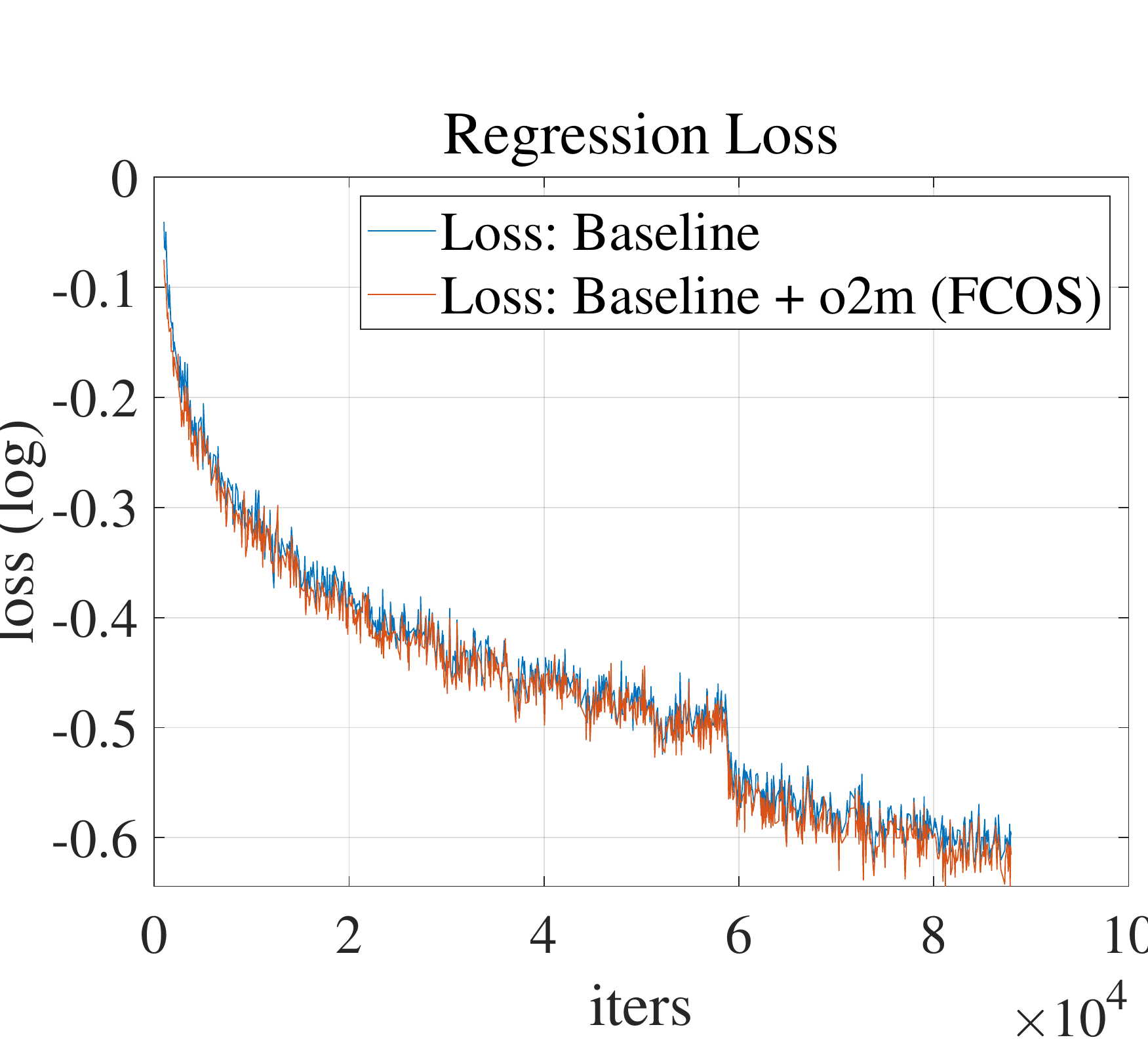}
        \caption{The regression loss of the o2o branch. }
        \label{fig:results-dual-assignment-reg-loss}
     \end{subfigure}
     \caption{The one-to-many assignment branch speeds up the learning of classification of the one-to-one assignment branch. }
     \label{fig:results-dual-assignment-loss}
\end{figure*}

\noindent \textbf{Basic Observations.} As shown in \cref{tab:results-observation}, compared with OneNet (trained with the one-to-one assignment policy), the detectors trained with the one-to-many assignment strategy converge faster, even if they have similar architecture. For example, after 12 epochs of training, RetinaNet and FCOS report 36.9 AP and 37.3 AP, respectively. In contrast, the performance of our baseline OneNet is 35.4 AP, which is much lower than RetinaNet and FCOS. These results imply that the one-to-many assignment might be significant in converging faster. We hypothesize that the sufficient positive samples helps the convergence of models with the one-to-many assignment. 

% Training longer (36 epochs) without changing the architecture of our baseline improves the performance, \ie, 38.6 AP. We guess that the convergence of OneNet is much slower. 

\begin{table*}[t!]
    \centering
    \begin{tabular}{l|c|c|rrrrrr}
        \toprule
        Model        & o2m & POTO 
        & $\text{AP}$ & $\text{AP}_{50}$ & $\text{AP}_{75}$ & $\text{AP}_s$ & $\text{AP}_m$ & $\text{AP}_l$ \\
        \midrule

        % 0.354 0.533 0.381 0.193 0.386 0.454
        Baseline (OneNet)   & - & \xmark 
        & 35.4 & 53.3 & 38.1 & 19.3 & 38.6 & 45.4 \\

        % 0.361 0.539 0.393 0.205 0.390 0.467
        Baseline (OneNet) & - & \cmark 
        & 36.1 & 53.9 & 39.3 & 20.5 & 39.0 & 46.7 \\

        \midrule

        % 0.368 0.546 0.399 0.201 0.398 0.483
        DATE-R (Ours) & RetinaNet & \xmark 
        & 36.8 & 54.6 & 39.9 & 20.1 & 39.8 & 48.3 \\

        % 0.370 0.549 0.404 0.205 0.398 0.490
        DATE-R (Ours)  & RetinaNet & \cmark 
        & 37.0 & 54.9 & 40.4 & 20.5 & 39.8 & \textbf{49.0} \\
        
        % 0.368 0.549 0.401 0.207 0.401 0.479
        DATE-F (Ours)  & FCOS & \xmark 
        & 36.8 & 54.9 & 40.1 & 20.7 & 40.1 & 47.9 \\

        % 0.373 0.553 0.407 0.212 0.403 0.488
        DATE-F (Ours) & FCOS & \cmark 
        & \textbf{37.3} & \textbf{55.3} & \textbf{40.7} & \textbf{21.2} & \textbf{40.3} & 48.8 \\

        \bottomrule

    \end{tabular}
    \caption{We study the impact of POTO and our Dual Assignment. We train all models for 12 epochs. }
    \label{tab:results-ablation}
\end{table*}

\begin{table*}[t!]
    \centering
    \begin{tabular}{l|c|llllll}
        \toprule
        Model        & epoch 
        & $\text{AP}$ & $\text{AP}_{50}$ & $\text{AP}_{75}$ & $\text{AP}_s$ & $\text{AP}_m$ & $\text{AP}_l$ \\
        \midrule
        
        % 0.373 0.553 0.407 0.212 0.403 0.488
        DATE-F & 12 
        & 37.3 (37.4) & 55.3 (57.0) & 40.7 (40.0) & 21.2 (22.0) & 40.3 (40.3) & 48.8 (48.4) \\

        % \midrule

        % 0.406 0.589 0.444 0.256 0.441 0.509 (w.o. NMS)
        DATE-F & 36 
        & 40.6 (40.5) & 58.9 (60.7) & 44.4 (43.3) & 25.6 (25.2) & 44.1 (43.9) & 50.9 (50.9) \\
        
        \bottomrule

    \end{tabular}
    \caption{The introduced one-to-many assignment doesn't influence the NMS-free property of the one-to-one assignment branch. Numbers in brackets are the results with NMS.}
    \label{tab:results-nms}
\end{table*}

\noindent \textbf{Hypothesis.} We suppose the one-to-many assignment helps the learning of the shared feature extractor and further speeds up the convergence of the one-to-one assignment branch. 
% We suppose the shared feature extractor might produce features suitable for both tasks, \ie, the one-to-one assignment and the one-to-many assignment. 

\noindent \textbf{On the Well-trained Feature Extractor.} We first design experiments to validate the importance of the feature extractor supervised by the one-to-many assignment. We initialize the multi-scale feature extractor of DATE with a trained FCOS. The one-to-one assignment branch remains randomly initialized. We set the max iterations to be 30000 iterations. The other settings remain the same with training DATE from scratch, \eg, training both branches together.

Results in \cref{fig:transferred} suggest that only about one-third of training time is enough to surpass the plain OneNet, \ie, 36.5 AP compared with 35.4 AP of OneNet. We attribute this phenomenon to the well initialized feature extractor. In our case, the trained feature extractor is suitable for the one-to-many assignment branch. The desired features for the one-to-one assignment branch might be similar to the one-to-many assignment branch, and we only need to do some tuning. 

We further remove the one-to-many assignment from our DATE and only train the one-to-one assignment branch. Results suggest that the one-to-many assignment branch has limited influence on the well-trained feature extractor, \ie, final performance dropped to 36.3 AP (0.2 AP). 

These phenomena indicate that a well-trained feature extractor is significant for detectors trained with the one-to-one assignment, regardless of whether there is a one-to-many assignment branch. However, the process of learning from scratch of such models is slow, \eg, OneNet. We suppose that the difference lies in the number of positive samples. These observations motivate us to propose our Dual Assignment. By introducing a branch trained with the one-to-many assignment strategy, our Dual Assignment provides more positive samples to supervise the learning of the feature extractor. 

\noindent \textbf{On the Introduction of Dual Assignment.} The introduced one-to-many assignment makes the one-to-one assignment branch converges faster. 
{The empirical evidence in \cref{fig:results-dual-assignment-loss} shows that the one-to-many assignment branch does speed up the convergence of the one-to-one assignment branch mainly by reducing the classification loss. We suppose the shared multi-scale feature extractor plays a significant role. The one-to-many assignment generates more positive samples than the one-to-many assignment branch, which might speed up the convergence of the feature extractor by reducing the necessity for more iterations of training. 
We guess converging earlier makes the feature extractor produce ``general'' features for classification. So the one-to-one assignment branch may mainly tune its parameters to fit the representations.  }
% We guess the shared parameters act like a shared term in the two objective functions, \ie, $l_{o2o}$ and $l_{o2m}$, and might dominate the values of functions. 

% \noindent \textbf{Introducing Dual Assignment.} 
% To enjoy the faster convergence speed, we bring back the one-to-many assignment branch $B_{o2m}$ alongside the one-to-one assignment branch $B_{o2o}$. 

\subsection{Ablation Studies}

\noindent \textbf{Dual Assignment and POTO.} By simply incorporating the one-to-many assignment branch, we observe non-trivial improvement on our baseline. As shown in \cref{tab:results-ablation}, when adopting the predictor (consisting of two convolutional layers) of RetinaNet as the one-to-many assignment branch, our Dual Assignment improves the performance from 35.4 AP to 36.8 AP, with a relative increase of 1.4 AP. A similar phenomenon happens when adopting FCOS as the one-to-many branch, \ie, DATE (36.8 AP) \vs OneNet (35.4 AP). 

OneNet adopts the cost of DETR~\cite{DETR-2020} as the quality measurement, while we prefer POTO~\cite{DeFCN-2021}. In \cref{tab:results-ablation}, POTO brings 0.7 AP improvement on the baseline OneNet. Adopting OneNet with POTO as the quality measurement, our DATE powered by FCOS also brings an increase of 1.2 AP, \ie, from 36.1 AP to 37.3 AP. 

Our Dual Assignment brings non-trivial improvement in the baseline, regardless of whether POTO is used as the quality measurement of Hungarian matching. These results indicate our Dual Assignment is significant for end-to-end detectors trained with the one-to-one assignment. 

% \noindent \textbf{Full Version DATE.}

\noindent \textbf{About NMS.} Introducing the one-to-many assignment branch does not influence the NMS-free property of our DATE. As shown in \cref{tab:results-nms}, even introducing the one-to-many assignment branch, the one-to-one assignment branch still performs end-to-end detection. For instance, the AP of our DATE is 37.3 and 37.4 when inference without and with NMS, respectively. We observe a similar phenomenon after 36 epochs of training, \eg, 40.6 AP (without NMS) \vs 40.5 AP (with NMS). This phenomenon suggests that the one-to-many assignment does not directly affect the prediction results of the one-to-one assignment branch. 
% It might be empirical evidence that the one-to-one assignment branch tunes the representations from the shared feature extractor. 

\begin{table}[t!]
    \centering
    \begin{tabular}{l|c|rrrrr}
        \toprule
        Model       & $\lambda_{o2m}$
        & $\text{AP}$ & $\text{AP}_{50}$ & $\text{AP}_s$ & $\text{AP}_m$ & $\text{AP}_l$ \\

        \midrule

        % 0.363 0.559 0.382 0.217 0.394 0.481
        FCOS & 1
        & 36.3 & 55.9 & \textbf{21.7} & 39.4 & 48.1 \\

        % 0.370 0.565 0.394 0.208 0.398 0.495
        FCOS & 4
        & \textbf{37.0} & \textbf{56.5} & 20.8 & \textbf{39.8} & \textbf{49.5} \\

        \midrule

        % 0.373 0.553 0.407 0.212 0.403 0.488
        DATE-F  & 1
        & \textbf{37.3} & \textbf{55.3} & \textbf{21.2} & \textbf{40.3} & \textbf{48.8} \\

        % 0.368 0.548 0.401 0.207 0.398 0.486
        DATE-F & 4
        & 36.8 & 54.8 & 20.7 & 39.8 & 48.6 \\

        \bottomrule
    \end{tabular}
    \caption{The influence of the weights of losses. FCOS: trained DATE-F but keeps the one-to-many instead of the one-to-one assignment branch. }
    \label{tab:results-preferences}
\end{table}
\noindent \textbf{Weights of Losses.} The weight of loss in a branch reflects the model's preferences, as shown in \cref{tab:results-preferences}. FCOS refers to the trained DATE-F but discards the one-to-one assignment branch instead of the one-to-many assignment branch. FCOS trained with larger weights reports better results among all metrics, \eg, AP increased from 36.3 to 37.0. In contrast, the performance of DATE dropped from 37.3 AP to 36.8 AP. 

% \subsection{Training Cost}

\begin{table}[t!]
    \centering
    \begin{tabular}{l|c|c|r|r}
    \toprule
        Model               & o2m  & phase & Params & FLOPs \\
    \midrule
        OneNet              & -    & infer & 32.0 M & 201   \\
    \midrule
        DATE-F (Ours)         & -    & infer & 32.0 M & 201   \\
        DATE-F (Ours)         & FCOS & train & 32.2 M & 205   \\
    \bottomrule
    \end{tabular}
    \caption{The extra computing resources and parameters during training are negligible. }
    \label{tab:results-flops-train}
\end{table}

\noindent \textbf{Training Cost.} The extra cost during training is negligible. For instance, as shown in \cref{tab:results-flops-train}, our DATE only adds 0.2M parameters (three layers of FCOS), which is 0.625\% of 32M of the original OneNet. The GFLOPs become 205 by an increment of 4 GFLOPs, which is 1.99\% of 201 GFLOPs of the original OneNet. Our Dual Assignment is friendly to machines with limited resources due to the negligible extra training cost and no inference overhead.

% \subsection{Exploration}\label{sec:results-sharing-parameters}

\begin{table}[t!]
    \centering
    \begin{tabular}{l|c|c|rr|rr}
        \toprule
        \multirow{2}{4em}{Model} & \multirow{2}{2.2em}{share subnet} & \multirow{2}{1em}{AP} & \multicolumn{2}{c}{Params} & \multicolumn{2}{|c}{GFLOPs} \\
        & & & train & infer & train & infer \\
        % & $\text{AP}$ & $\text{AP}_{50}$ & $\text{AP}_{75}$ & $\text{AP}_s$ & $\text{AP}_m$ & $\text{AP}_l$ \\
        \midrule

        % 0.369 0.567 0.390 0.207 0.402 0.498
        RetinaNet
        & - & 36.9 & 33.6 & 33.6 & 233 & 233 \\
        
        % 0.370 0.549 0.404 0.205 0.398 0.490
        DATE-R
        & \cmark & 37.0 & 33.8 & 32.0 & 238 & 201 \\

        % 0.376 0.556 0.407 0.215 0.406 0.500
        DATE-R
        & \xmark & \textbf{37.6} & 38.5 & 32.0 & 388 & 201 \\

        \bottomrule
    \end{tabular}
    \caption{Not sharing subnets increases the performance of DATE (RetinaNet-style one-to-many assignment branch).}
    \label{tab:results-sharing}
\end{table}

% \noindent \textbf{Sharing How Many Parameters.}
\noindent \textbf{Not Sharing Subnets.} As described in \cref{sec:method}, we share the parameters of most layers, including classification and regression subnets. 
% There are only parameters of two or three convolutional layers not shared. 
This design brings almost no extra cost during training. A new question arises: what happens if we make the two subnets not shared? We conduct experiments and present results in \cref{tab:results-sharing}. We adopt two convolutional layers of RetinaNet to construct the one-to-many branch. Making the two subnets separate suggests that this modification can improve the final performance but introduces more parameters and FLOPs during training. For example, when adopting RetinaNet output layers to construct the one-to-many assignment branch, the total parameters during training of our DATE are 33.8 M (the output channels of RetinaNet are nine times more than FCOS, \ie, 1.8 M \vs 0.2 M). Making subnets separate for the two branches increases the AP of DATE from 37.0 to 37.6. However, this modification introduces about 4.7M parameters, which is 14\% of the original DATE (during training). Besides, the total FLOPs increased by about 42\%, from 238G to 338G.

\section{Conclusion}
In this paper, we propose a Dual Assignment to solve the slow convergence issue of end-to-end fully convolutional object detection (DATE). We show our method is orthogonal to network modification and does not cost extra computing resources during inference. The negligible extra training cost makes our DATE friendly to researchers. These features enable our DATE to serve as a strong baseline. 

%%%%%%%%% REFERENCES
{\small
\bibliographystyle{ieee_fullname}
\bibliography{egbib}
}

\appendix

\begin{table*}[!ht]
    \centering
    \begin{tabular}{c|cc|cccccc}
    \toprule
        Model & $\lambda_{o2o}$ & $\lambda_{o2m}$ 
        & $\text{AP}$ & $\text{AP}_{50}$ & $\text{AP}_{75}$ & $\text{AP}_s$ & $\text{AP}_m$ & $\text{AP}_l$ \\
        
        \midrule
        
        % 0.354 0.533 0.381 0.193 0.386 0.454
        Baseline (OneNet)  & - & - 
        & 35.4 & 53.3 & 38.1 & 19.3 & 38.6 & 45.4 \\
        
        % % 0.361 0.539 0.393 0.205 0.390 0.467
        % Baseline (OneNet) & - & \cmark 
        % & 36.1 & 53.9 & 39.3 & 20.5 & 39.0 & 46.7 \\
        
        \midrule
        
        \multirow{7}{5em}{DATE-F} 
        
        % 0.372 0.553 0.408 0.210 0.402 0.490
        & 1 & 0.5 
        & 37.2 & 55.3 & 40.8 & 21.0 & 40.2 & 49.0 \\
        
        % 0.373 0.553 0.407 0.212 0.403 0.488
        & 1 & 1 
        & \cellcolor{Gray}{\underline{37.3}} & \cellcolor{Gray}{55.3} & \cellcolor{Gray}{40.7} & \cellcolor{Gray}{21.2} & \cellcolor{Gray}{40.3} & \cellcolor{Gray}{48.8} \\
        
        % 0.372 0.548 0.408 0.209 0.401 0.488
        & 1 & 2 
        & 37.2 & 54.8 & 40.8 & 20.9 & 40.1 & 48.8 \\
        
        % configs/e2edet/e2edet_fcos_devA02_r50_12e_8x2_fcos_poto_dev09.py
        % 0.368 0.548 0.401 0.207 0.398 0.486
        & 1 & 4 
        & 36.8 & 54.8 & 40.1 & 20.7 & 39.8 & 48.6 \\
        
        % 0.370 0.550 0.402 0.216 0.401 0.489
        & 0.5 & 1 
        & 37.0 & 55.0 & 40.2 & 21.6 & 40.1 & 48.9 \\
        
        % 0.375 0.555 0.411 0.224 0.404 0.491
        & 2 & 1 
        & \textbf{37.5} & \textbf{55.5} & \textbf{41.1} & \textbf{22.4} & \textbf{40.4} & \textbf{49.1} \\
        
        % 0.368 0.545 0.398 0.215 0.400 0.482
        & 4 & 1 
        & 36.8 & 54.5 & 39.8 & 21.5 & 40.0 & 48.2 \\
        
        \midrule
        
        \multirow{7}{5em}{DATE-R} 
        
        % 0.369 0.549 0.400 0.200 0.399 0.485
        & 1 & 0.5 
        & 36.9 & 54.9 & 40.0 & 20.0 & 39.9 & 48.5 \\
        
        % 0.372 0.552 0.404 0.199 0.404 0.486
        & 1 & 1 
        & \textbf{37.2} & 55.2 & \textbf{40.4} & 19.9 & 40.4 & 48.6 \\
        
        % 0.370 0.549 0.404 0.205 0.398 0.490
        & 1 & 2 
        & \cellcolor{Gray}{\underline{37.0}} & \cellcolor{Gray}{54.9} & \cellcolor{Gray}{\textbf{40.4}} & \cellcolor{Gray}{20.5} & \cellcolor{Gray}{39.8} & \cellcolor{Gray}{\textbf{49.0}} \\
        
        % 0.367 0.546 0.397 0.197 0.400 0.489
        & 1 & 4 
        & 36.7 & 54.6 & 39.7 & 19.7 & 40.0 & 48.9 \\
        
        % 0.371 0.553 0.402 0.215 0.405 0.489
        & 0.5 & 1 
        & \underline{37.1} & \textbf{55.3} & 40.2 & \textbf{21.5} & \textbf{40.5} & 48.9 \\
        
        % 0.368 0.549 0.397 0.203 0.399 0.485
        & 2 & 1 
        & 36.8 & 54.9 & 39.7 & 20.3 & 39.9 & 48.5 \\
        
        % 0.362 0.541 0.395 0.198 0.395 0.476
        & 4 & 1 
        & 36.2 & 54.1 & 39.5 & 19.8 & 39.5 & 47.6 \\

    \bottomrule        
    \end{tabular}
    \caption{Ablation studies on the weights of the one-to-one and the one-to-many assignment branches. We don't carefully tune the weights in our main results. The bold numbers are the best results. Numbers in the rows with gray as the background color result from default settings. We set $\lambda_{o2m}$ of DATE-R as 2 in our main results to balance the loss. Settings that are 0.2AP worse than the best result are underlined.}
    \label{tab:results-lambda}
\end{table*}

\section{Additional Experimental Results}
\subsection{Weights of Losses}
The weights of losses to produce our main results are not carefully designed. Thus we present the results with different weights in \cref{tab:results-lambda}. Results indicate that simply setting $\lambda_{o2o}$ and $\lambda_{o2m}$ as 1 is sufficient to produce good enough results. This property is favorable as it saves the efforts of researchers by avoiding engineering design. 
% the our DATE can further be improved by selecting more proper weights. For example, setting $\lambda_{o2o} = 2$ and $\lambda_{o2m}=1$ gives DATE-F an increase of 0.2 AP, \ie, from 37.3 to 37.5. 

\subsection{On Initializing Feature Extractor from Trained Models}

\begin{table}[t!]
    \centering
    \begin{tabular}{c|cccccc}
        \toprule
        
        Model 
        & $\text{AP}$ & $\text{AP}_{50}$ & $\text{AP}_{75}$ & $\text{AP}_s$ & $\text{AP}_m$ & $\text{AP}_l$ \\
        
        \midrule
        
        OneNet & 
        38.6 & 56.6 & 42.2 & 23.9 & 41.7 & 48.7 \\
        
        % 0.402 0.585 0.440 0.243 0.438 0.506
        OneNet$\dagger$ 
        & 40.2 & 58.5 & 44.0 & 24.3 & 43.8 & 50.6 \\
        
        DATE-F 
        & \textbf{40.6} & \textbf{58.9} & \textbf{44.4} & \textbf{25.6} & \textbf{44.1} & \textbf{50.9} \\
        
        \bottomrule
    \end{tabular}
    \caption{OneNet with the feature extractor initialized from a well-trained FCOS is still inferior to our DATE-F. $\dagger$ means the parameters of the feature extractor is initialized from a trained FCOS. }
    \label{tab:results-feature-extractor}
\end{table}

One may suppose that a well-trained feature extractor is sufficient to train a OneNet with good performance, \eg, initializing the parameters of the feature extractor (backbone+FPN+subnets) of OneNet from a trained FCOS. We abbreviate it as \textit{Initialization Strategy}. Here we point out the two shortcomings of this strategy: 
\begin{enumerate}
    \item Performance: the final performance is still inferior to our DATE.
    \item Training cost: the training cost is double of that of our DATE.
\end{enumerate}
We will discuss them in the following.

\noindent \textbf{Implementation Details.} 
% We adopt the 3x training setting with multi-scale data augmentation. \textcolor{red}{The 1x setting without multi-scale augmentation may lead to the overfitting of models with the feature extractor initialized from a trained network. } 
We initialize the parameters of the feature extractor of models marked by $\dagger$ with a trained FCOS (3x training setting). Other settings remain unchanged.

\noindent \textbf{Performance.} As shown in \cref{tab:results-feature-extractor}, the OneNet with feature extractor initialized from a well-trained FCOS, marked by $\dagger$, reports 40.2 AP, which is still inferior to our DATE-F (40.6 AP). It should be emphasized that our DATE-F does not adopt the well-trained feature extractor but still outperforms OneNet$\dagger$. This result implies that our Dual Assignment is significant in helping the improvement of the final performance.

% \begin{table}[t!]
%     \centering
%     \begin{tabular}{c|rr}
%         \toprule
%         Module & Params (M) & GFLOPs \\
%         \midrule
%         Feature Extractor & 31.8 & 197\\
%         Cls \& Reg Layers & 0.2 & 4 \\
%         \midrule
%         Total             & 32.0 & 201 \\
%         \bottomrule
%     \end{tabular}
%     \caption{Parameters and FLOPs of OneNet. }
%     \label{tab:results-resources-onenet}
% \end{table}

\noindent \textbf{Training Cost.} The Initialization Strategy, \ie, training a OneNet with its feature extractor initialized from a well-trained model, almost doubles the training cost. This strategy requires training two models: 1) Training a model with the same feature extractor using the one-to-many assignment; 2) Training a model with the feature extractor initialized from 1) using the one-to-one assignment. In contrast, we only train one model with negligible additional cost (Section 4.7). Once the architecture changed, \eg, a stronger backbone or replacing FPN~\cite{FPN-2017} with BiFPN~\cite{BiFPN-2020}, it's inevitable for the Initialization Strategy to repeat the above steps. 

\section{Brief Introduction to Pareto Optimal} \label{sec:pareto}
Here we briefly introduce Pareto optimality~\cite{Pareto-1906}, and we refer readers to other materials~\cite{ParetoOptimality-2017}. 

A typical multi-objective optimization problem is to minimize: 
\begin{align}
    \mathbf{f}(\mathbf{x}) = \begin{bmatrix} 
        f_1(\mathbf{x}), f_2(\mathbf{x}), \ldots, f_m(\mathbf{x})
    \end{bmatrix}^\intercal
\end{align}
subject to: 
\begin{equation}
    \begin{split}
        c_i(\mathbf{x}) = 0,    & \quad i \in \mathcal{E} \\
        c_j(\mathbf{x}) \le 0,  & \quad j \in \mathcal{I}
    \end{split}
\end{equation}
where $\mathbf{x} \in \mathbb{R}^n$, $m$ is the total number of objective functions, $\mathcal{E}, \mathcal{I}$ are the finite sets of indices of equality and inequality constraints, respectively. The \textit{feasible set} $\mathcal{S}$ (also called \textit{feasible design space}) is defined as a collection of all the feasible points $\mathbf{x}$ that satisfy the constraints, that is: 
\begin{equation}
    \mathcal{S} = \{\mathbf{x}|c_i(\mathbf{x})=0, i \in \mathcal{E}; c_j(\mathbf{x}) \le 0, j \in \mathcal{I}\}
\end{equation}
so the formulation is simplified as:
\begin{equation}
    \min_{\mathbf{x} \in \mathcal{S}} \mathbf{f}(\mathbf{x})
\end{equation}
We also define the \textit{feasible criterion space} $\mathcal{Z}$ as:
\begin{equation}
    \mathcal{Z} = \{\mathbf{f}(\mathbf{x})|\mathbf{x} \in \mathcal{S}\}
\end{equation}

Ideally, we would like to seek a point $\mathbf{x}^u$ that minimizes all objective functions. 
\begin{definition}[Utopia Point]\label{def:utopia-point}
A point $\mathbf{f}^u$ in the criterion space is called the utopia point if $f^u_i = \min f_i(\mathbf{x}), \forall \mathbf{x} \in \mathcal{S}$.
\end{definition}

\begin{definition}[Pareto Optimality]
A point $\mathbf{x}^P$ in the feasible set $\mathcal{S}$ is Pareto optimal if and only if: $\nexists \mathbf{x} \in \mathcal{S}$ that satisfying $\mathbf{f}(\mathbf{x}) \le \mathbf{f}(\mathbf{x}^P)$ with at least one $f_i(\mathbf{x}) < f_i(\mathbf{x}^P)$.
\end{definition}

\begin{figure}[t!]
    \centering
    \includegraphics[trim={2cm 6.5cm 2cm 7cm}, clip, width=0.45\textwidth]{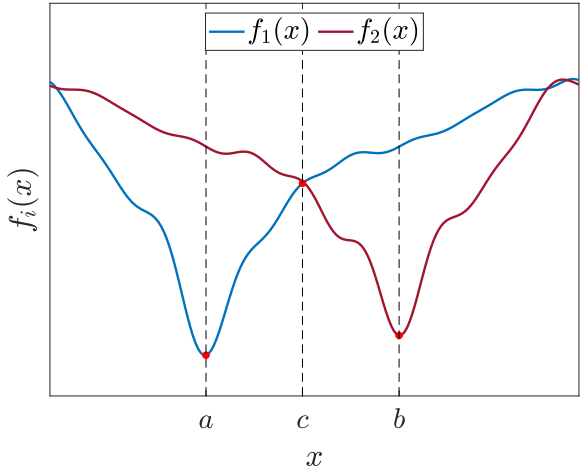}
    \caption{An example of Pareto optimal.}
    \label{fig:pareto-optimal}
\end{figure}

\cref{fig:pareto-optimal} is an example of Pareto optimal. By definition, we can easily recognize that the solution $x=c$ is Pareto optimal. Actually, solutions in the closed interval $[a, b]$ are all Pareto optimal. The set of all Pareto optimal points is called the \textit{Pareto front} (or \textit{Pareto frontier}).

There are many methods for multi-objective optimization. One of the most common approaches is the \textit{weighted sum} method:
\begin{equation}
    g(\mathbf{x}) = \mathbf{w}^\intercal \cdot \mathbf{f}(\mathbf{x}) = \sum_{i=1}^m w_i f_i(\mathbf{x})
\end{equation}
where $\mathbf{w}^\intercal=[w_1, w_2, \ldots, w_m]$ are the weights of their corresponding objective functions. Intuitively, the weights would reflect the preferences of the user. A larger weight would make the optimizer pay more attention to the corresponding objective function. In an extreme case, if we set all but $w_t$ as zero, the multi-objective optimization degrades into a single-objective optimization. The optimizer minimizes $f_t(x)$ without regard for other objective functions.

\end{document}